\documentclass[letterpaper, 10 pt, conference]{ieeeconf}  % Comment this line out if you need a4paper

\IEEEoverridecommandlockouts                              % This command is only needed if 
\title{\LARGE \bf
Contact-Aware Imitation Learning Through Contact Factorization}

\author{Jiho Hong$^{1}$, Daeun Song$^{2*}$, Sanghyun Kim$^{1*}$, Mingyo Seo$^{3*}$
\thanks{$^{1}$Kyung Hee University, $^{2}$Ewha Womans University, $^{3}$University of Central Florida.}
\thanks{$^{*}$Correspondence: {\tt\small songd@ewha.ac.kr}, {\tt\small kim87@khu.ac.kr}, {\tt\small mingyo.seo@ucf.edu}} 
}

\let\labelindent\relax
\DeclareUnicodeCharacter{0301}{\'{e}}
\usepackage{times}
\usepackage{xcolor}
\usepackage{amsfonts}
\usepackage{amsmath}
\usepackage{amssymb}
\usepackage{multirow}
\usepackage{colortbl}
\usepackage{cuted}
\usepackage{capt-of}
\usepackage{graphicx}
\usepackage{grffile}
\usepackage{booktabs}
\usepackage{hyperref}
\usepackage{dsfont}
\usepackage{svg}
\usepackage{setspace}
\usepackage[font=scriptsize, labelfont=bf]{caption}
\usepackage{enumitem}
\usepackage[backend=biber,
            hyperref=true,
            url=false,
            isbn=false,
            doi=false,
            backref=false,
            style=ieee,
            citestyle=numeric-comp,
            sorting=none,
            block=none]{biblatex}
\usepackage[font=footnotesize]{caption}

\usepackage{algpseudocode}
\usepackage{algorithm}
\usepackage{algpseudocode}
\usepackage{dblfloatfix}
\usepackage[normalem]{ulem}
\usepackage{booktabs}
\usepackage{makecell}
\usepackage{pifont}

\usepackage{bm} %%
\usepackage{tikz}
\usetikzlibrary{arrows.meta,positioning,calc}

\setlist[itemize]{leftmargin=*}
\setlist[enumerate]{leftmargin=*}

\renewcommand{\bibfont}{\small}
\newcommand{\ourmethod}{{\textsc{FACE}}}
\newcommand{\failrates}[4]{%
  (\textcolor{purple}{#1\%}, \textcolor{orange}{#2\%},
   \textcolor{green!50!black}{#3\%}, \textcolor{red}{#4\%})%
}

\algblock{ParFor}{EndParFor}
\algnewcommand\algorithmicparfor{\textbf{parallel for}}
\algnewcommand\algorithmicpardo{\textbf{do}}
\algnewcommand\algorithmicendparfor{\textbf{end\ parallel}}
\algrenewtext{ParFor}[1]{\algorithmicparfor\ #1\ \algorithmicpardo}
\algrenewtext{EndParFor}{\algorithmicendparfor}
\begin{document}
\maketitle
\thispagestyle{empty}
\pagestyle{empty}

%===============================================================================

\begin{abstract}
Generalizable contact-rich manipulation requires robots to preserve intended task behavior while adapting its physical realization to changing contact conditions.
However, interaction forces can vary substantially with small changes in surface geometry, orientation, and friction, making policies trained directly on raw force measurements difficult to transfer beyond demonstrated conditions.
We introduce \ourmethod{}, a contact-factorized imitation learning framework that separates intended task behavior from environment-dependent contact factors.
Our representation expresses interaction forces in normalized, contact-relative coordinates, while a learned contact-normal estimator and an online friction estimator infer the local contact normal and effective friction scale.
Together, these estimators enable force observations to be encoded and policy outputs to be decoded into physical motion and force commands during execution.
In this way, \ourmethod{} adapts execution to current contact conditions while preserving the intended task behavior, without updating the policy parameters. 
We evaluate \ourmethod{} on real-robot contact-rich manipulation under unseen variations in surface properties and geometry, demonstrating robust generalization across contact conditions through controlled comparisons with variants that adapt prior approaches to our setting. 
Videos and additional materials can be found on the project page: \url{https://rcilab.khu.ac.kr/face}.
\end{abstract}
%===============================================================================

\section{Introduction}

Recent advances in robot learning have enabled robots to acquire complex manipulation behaviors from scalable human demonstrations~\cite{mandlekar2021matters, zhu2022viola, chi2024diffusionpolicy, Ze2024DP3}. This progress has been strong for tasks in which visual observations provide much of the information needed to choose an action. However, contact-rich manipulation also depends on physical properties of the environment that may not be known before interaction. The same intended behavior can require different forces across surfaces or objects, and those forces may need to be applied along different directions as the contact geometry changes. Consequently, directly imitating raw force measurements can tie a learned policy to the conditions seen during demonstration rather than the interaction required by the task.

%Recent work has incorporated force/torque or tactile feedback into visuomotor policies and has learned force commands, compliance profiles, or high-frequency corrective actions for contact-rich manipulation~\cite{yu2024mimictouch, liu2025forcemimic, choi2026umift, liu2025factr}. These approaches substantially improve contact awareness and robustness. However, many learning-based methods still represent interaction forces directly in physical quantities whose scale and orientation vary with the environment. For sliding contact, for example, the required tangential force changes with friction and normal loading, while the relevant force directions change with the surface orientation. As a result, generalization across contact conditions must either be covered by the demonstration distribution or learned implicitly by the policy.

Force/torque and tactile feedback have enabled force-aware imitation learning, tactile-guided control, adaptive compliance, and multimodal policy learning for contact-rich manipulation~\cite{yu2024mimictouch, liu2025forcemimic, choi2026umift, liu2025factr, jones24fuse,yu2025forcevla,li2026atvla}. These methods substantially improve contact awareness and robustness, but they generally represent forces in physical quantities whose scale and orientation depend on the environment. During sliding, the required tangential force varies with friction and normal loading, while force directions vary with surface orientation. Consequently, generalization across contact conditions can depend heavily on demonstration coverage or adaptation learned implicitly by the policy.

\begin{figure}[t]
    \centering
    \includegraphics[width=1\columnwidth]{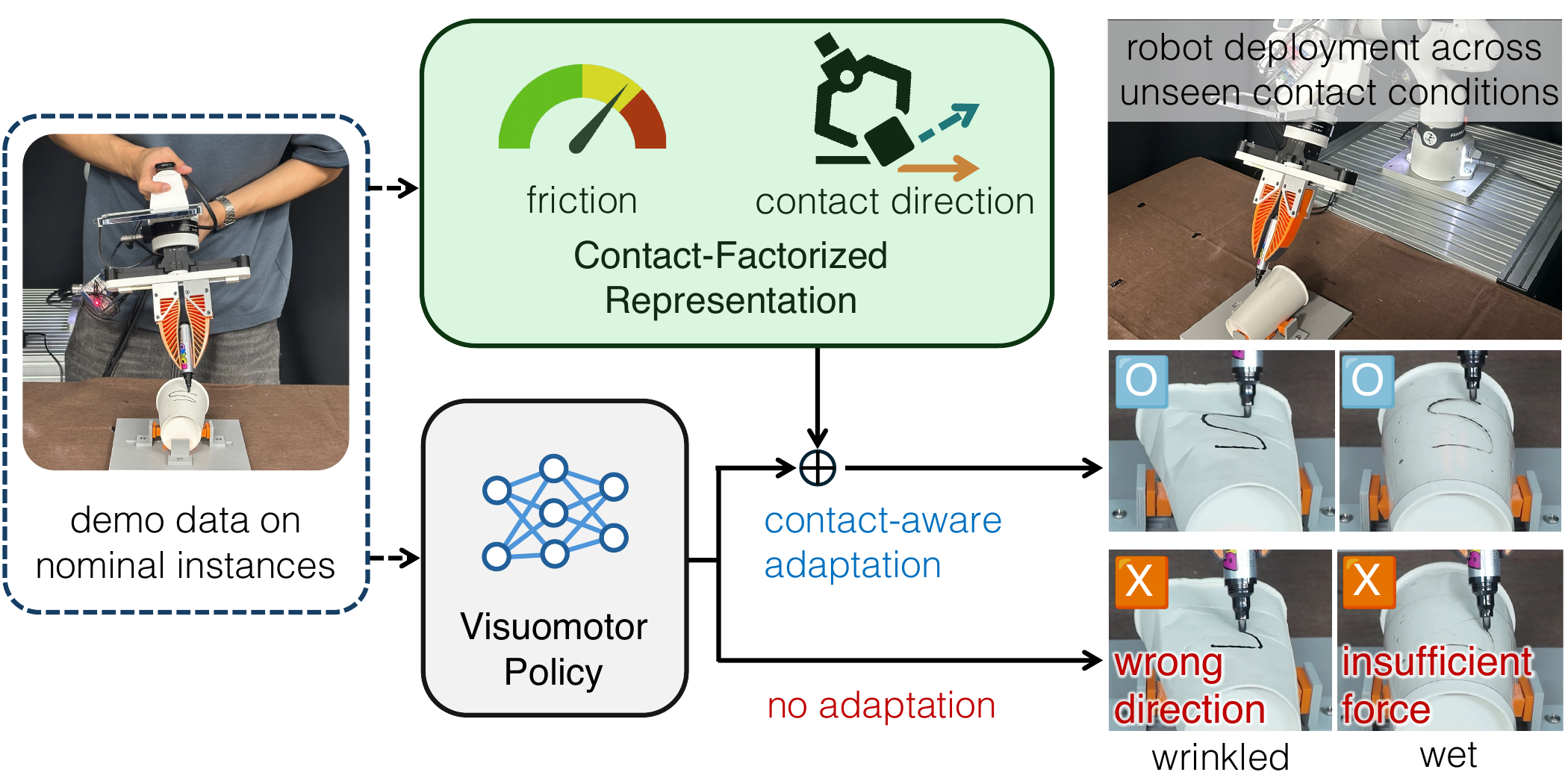}
    \caption{
\textbf{Overview of \ourmethod{}.}
\ourmethod{} addresses contact-rich manipulation under uncertain physical conditions by separating the demonstrated interaction from contact-dependent force magnitude and direction. A contact-normal estimator trained through interactions with nominal object instances is combined with an online friction estimator during execution.
This adaptation adjusts interaction forces online while preserving the intended behavior.
}
    \label{fig:overview}
\end{figure}

Our key idea is to separate what is task-specific from what changes with the contact environment. The intended task behavior should be preserved, while its physical motion and force commands adapt to the current contact conditions. We realize this idea with normalized force coordinates defined in the local contact frame: normal loading is expressed relative to prescribed operating bounds, while tangential loading is expressed relative to friction-scaled normal loading. The same contact factorization is used for observations and actions, while friction scale and contact orientation are estimated separately and used both to encode force observations and to map the learned output back to physical commands.

To capture the intended task behavior, we collect demonstrations with a handheld gripper equipped with force/torque sensing and visual-inertial tracking. The handheld form allows the demonstrator to physically perform contact-rich interactions while synchronized end-effector motion and interaction wrench are recorded. At deployment, the friction scale and local contact normal are estimated online and used to map normalized policy outputs back to physical motion and force commands. The policy itself remains fixed, allowing the demonstrated behavior to be preserved across surfaces with different frictional properties and curvatures.

\begin{figure*}[t]
    \centering
    \vspace{4pt}
    \includegraphics[width=1.8\columnwidth]{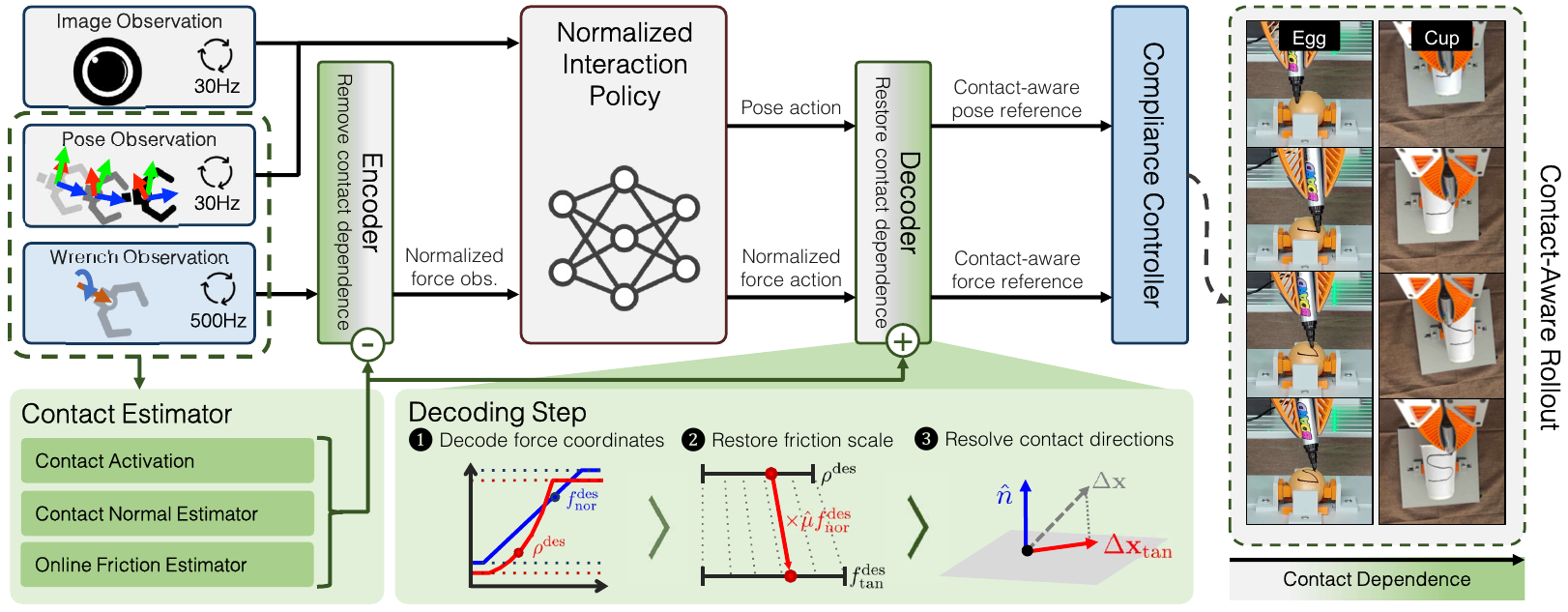}
    \caption{\textbf{Model architecture and contact-aware online adaptation.} Visual and wrench observations are encoded into normalized interaction coordinates that remove contact-dependent force variation. The policy predicts normalized pose and force actions, which are decoded using online estimates of contact activation, surface normal, and friction to restore the appropriate force scale and direction. The resulting contact-aware references are tracked by a compliance controller, enabling robust execution across varying contact conditions.}
    \label{fig:system}
    \vspace{-0.3cm}
\end{figure*}

\iffalse
In summary, our contributions are threefold:
\begin{itemize}
\item A contact-factorized representation for imitation learning that expresses bounded normal loading and friction-relative tangential loading in the local contact frame, with corresponding normalized coordinates for both force observations and actions.
\item An online physical reparameterization framework that estimates environment-dependent friction and contact orientation and uses them to decode a fixed learned policy into condition-appropriate motion and force commands without policy adaptation.
\item A real-robot implementation and evaluation of the proposed framework, including a force-sensing handheld gripper for collecting contact-rich demonstrations, and systematic experiments that vary friction, surface geometry, and their combination to validate generalization under changing contact conditions.
\end{itemize}
\fi

In summary, our contributions are threefold:
\begin{itemize}
\item A contact-factorized representation of force observations and actions, using bounded normal loading and friction-relative tangential loading in the local contact frame.

\item An online reparameterization framework that converts outputs of a fixed policy into motion and force commands using the estimated friction scale and contact normal.

\item A force-sensing handheld demonstration interface and real-robot evaluation under variations in friction, surface geometry, and their combination.
\end{itemize}

\section{Related Work}
\label{sec:related}

% Contents to be written; citations deferred for now.

\subsection{Force-Aware Demonstration and Imitation Learning}

Recent work has increasingly incorporated physical interaction signals into imitation learning. One common approach acquires force-aware demonstrations through teleoperation and enables policies to attend explicitly to interaction feedback \cite{liu2025factr}. While effective, such systems require an instrumented human-robot interface and closed-loop robot execution during demonstration, and the recorded force signals remain tied to the physical conditions under which they are collected.

Complementary to teleoperation-based collection, robot-free and human-centered interfaces enable demonstration collection through handheld tools and direct human interaction \cite{song2020grasping,young2021visual, chi2024umi, seo2024legato,shafiullah2023bringing,yu2024mimictouch}. More recent systems extend this paradigm to capture interaction forces during demonstration and learn force-aware manipulation policies \cite{liu2025forcemimic,choi2026umift}. These approaches facilitate natural acquisition of coupled motion and force information and improve contact regulation. Their learned force and compliance targets can nevertheless remain dependent on the contact conditions represented in the demonstrations.

Beyond data collection, recent visuomotor policies incorporate force or tactile sensing through reactive architectures, multimodal fusion, and slow-fast policy structures \cite{he2025foar,xue2025reactive,chen2026implicitrdp}. Concurrent work also analyzes characteristic failure modes of force-aware vision-language-action policies in contact-rich tasks \cite{paresmorlans2026fact}. Together, these works establish the importance of physical interaction feedback. Our focus is on how that interaction is represented across changing contact conditions.

\subsection{Contact-Aware Action Representation}

Beyond force-aware observation and demonstration, another key question is how contact-rich actions should be represented and executed. 
Classical force-control methods structure contact-rich actions by separating force- and motion-controlled directions in task-related coordinates \cite{raibert1981hybrid, bruyninckx1996specification}, while later learning-based approaches infer changing constraint frames from demonstrations \cite{conkey2019learning}.
Learning-based work has also explored action representations that expose physically meaningful control variables, such as end-effector motion together with variable impedance, to improve learning in contact-rich tasks \cite{martin2019variable}.
More recent policies learn task-dependent compliance or force regulation \cite{hou2025adaptive,xu2025crdagger}, and combine learned visuomotor policies with interaction-frame estimation or hybrid force-position execution \cite{fang2026forcepolicy,li2026forcevla2}. These approaches improve contact stability and force regulation through learned compliance and contact-aware execution. We focus on normalizing tangential loading by friction-scaled normal loading to represent interaction across changing contact conditions.

Other methods use physical feedback at inference time to calibrate predicted states or steer generated actions toward safer or more feasible interactions \cite{wei2025safediff,zhang2026touchguide, wu2026vital}. Concurrent work further explores fast force reflexes, asynchronous vision-force policy composition, and object-centric contact-wrench guidance \cite{huang2026unireflex,he2026lagfusion,zhu2026chord}. Our approach uses the estimated contact normal and friction scale to encode force observations in contact-factorized coordinates and decode the outputs of a fixed policy into physical motion and force commands.

\section{Method}
\label{sec:method}

We introduce \ourmethod{} (\underline{F}orce \underline{A}daptation through \underline{C}ontact-Factor \underline{E}stimation), an imitation-learning framework for contact-rich manipulation using a contact-factorized representation.
The key idea is to represent environment-dependent physical interaction through a compact contact factor, separating it from the task behavior demonstrated by the user, as illustrated in Fig.~\ref{fig:overview}.
A visuomotor policy therefore learns receding-horizon actions in contact-factorized coordinates, while the contact factor is estimated by a separate module that requires no task-specific supervision.
At deployment, online contact-factor estimates transform the outputs of a fixed policy into pose and force references consistent with the current contact condition.

\subsection{Problem Formulation}
\label{sec:method:problem}

We formulate contact-rich manipulation of objects and surfaces with unknown physical properties and local contact geometry as a partially observed Markov decision process
\(\mathcal{M}=(\mathcal{S},\mathcal{O},\mathcal{A},P,R,\gamma)\), where $\mathcal{S}$ is the state space, $\mathcal{O}$ is the observation space, $\mathcal{A}$ is the action space, $\mathcal{P}(\cdot|s, a)$ is the transition probability, $R(s, a, s')$ is the reward function, and $\gamma \in [0, 1)$ is the discount factor.
At time \(t\), we decompose the latent state as \(s_t=(x_t,z_t)\), with the contact factor \(z_t=(\bm{n}_t,\mu_t)\), where \(x_t\) denotes the task state, \(\bm{n}_t\) the local contact normal, and \(\mu_t>0\) the effective friction scale.
The contact factor is generally unknown before interaction and may vary across objects, surfaces, and contact locations.

The robot receives the observation \(\bm{y}_t=(\bm{I}_t,\bm{p}_t,\bm{w}_t)\in\mathcal{O}\), where \(\bm{I}_t\), \(\bm{p}_t\), and \(\bm{w}_t\) denote the RGB observation, end-effector pose, and measured interaction wrench, respectively.
Although \(z_t\) is not directly observed, its effect is reflected in the history of tool motion and interaction wrench.
We therefore estimate the contact factor as \(\hat z_t=\phi(\bm{y}_{t-L:t})\) and use it to factorize the interaction quantities, separating the demonstrated behavior from its contact-dependent physical realization.

Given \(\hat z_t\), the raw interaction observation is mapped to the factorized coordinate \(\tilde{\bm{w}}_t=\mathcal{E}(\bm{w}_t;\hat z_t)\).
The visuomotor policy then predicts the receding-horizon sequence
\(\bm{a}_{t:t+H-1}=\pi(\bm{I}_t,\bm{p}_t,\tilde{\bm{w}}_t)\), with each action represented as
\(\bm{a}_t=(\Delta\bm{x}_t,\Delta\bm{R}_t,\tilde f_{\mathrm{nor},t},c_t)\).
The policy output is transformed using the current contact-factor estimate to obtain physical motion and force references.

\subsection{Contact Factorization for Interaction Representation}
\label{sec:method:representation}

%We seek an interaction representation that preserves the demonstrated contact behavior while factoring out quantities that depend on the local contact condition. We define the contact factor as \(z_t=(\bm{n}_t,\mu_t)\), where \(\bm{n}_t\) captures the local contact orientation and \(\mu_t\) captures the friction-dependent scale of tangential loading. Using \(z_t\), the same physical interaction can be expressed in contact-relative coordinates, while its realization under the current contact condition is handled separately.

Following UMI~\cite{chi2024umi}, we express pose observations and motion actions relative to the current end-effector pose.
This removes dependence on the absolute robot configuration and world frame, allowing the policy to represent motion locally with respect to the task.
All force quantities and the contact normal \(\bm{n}_t\) are expressed in the current end-effector frame.
Resolving the measured force with respect to \(\bm{n}_t\) then expresses interaction relative to the local surface rather than fixed end-effector axes.
We use a unit contact normal oriented so that compressive normal loading is positive, and obtain normal and tangential forces by projecting the measured force onto this normal and its orthogonal plane.

We treat normal loading as a task-level target specifying the intended level of contact.
We normalize it by a prescribed operating limit,
\begin{equation}
\tilde f_{\mathrm{nor},t}
=
\Pi_{[0,1]}
\left(
\frac{f_{\mathrm{nor},t}}{f_{\mathrm{nor},\max}}
\right),
\label{eq:normal_force_normalization}
\end{equation}
where \(f_{\mathrm{nor},\max}\) denotes the prescribed maximum normal loading and \(\Pi\) denotes clipping to the indicated interval.
The resulting coordinate is dimensionless and bounded, with zero corresponding to no normal loading and one to loading at or above the prescribed maximum.

Tangential loading, in contrast, depends strongly on the frictional properties of the contact and on the applied normal force.
For an effective friction scale \(\mu>0\), we therefore factorize the tangential force magnitude as
\begin{equation}
f_{\mathrm{tan},t}
=
\mu f_{\mathrm{nor},t}\rho_t(\mu),
\label{eq:tangential_factorization}
\end{equation}
where
\(
\rho_t(\mu)
=
f_{\mathrm{tan},t}/(\mu f_{\mathrm{nor},t})
\)
denotes the multiplicative residual after factoring out the contact-dependent scale \(\mu f_{\mathrm{nor},t}\).
The residual is evaluated only when \(f_{\mathrm{nor},t}>f_{\epsilon}\), where \(f_{\epsilon}>0\) avoids ill-conditioned normalization near zero normal loading.
Under steady Coulomb sliding, \(\rho_t(\mu)\) approaches unity when \(\mu\) matches the effective friction scale.
Thus, \(\rho_t\) represents the demonstrated tangential loading relative to the friction-scaled normal loading.

For the interaction component of the action, we represent this multiplicative residual in bounded log coordinates as
\begin{equation}
c_t(\mu)
=
\log
\Pi_{[\rho_{\min},\rho_{\max}]}
\left(
\rho_t(\mu)
\right).
\label{eq:tangential_force_coordinate}
\end{equation}
The logarithm converts multiplicative deviations around the nominal value \(\rho_t=1\) into additive deviations around \(c_t=0\), providing a more convenient coordinate for policy learning.
Here, \(\rho_{\min}\) and \(\rho_{\max}\) denote prescribed bounds used for action generation, satisfying \(0<\rho_{\min}\leq1\leq\rho_{\max}\).
Samples with \(f_{\mathrm{nor},t}\leq f_{\epsilon}\) are excluded from supervision of the tangential action coordinate.

The interaction observation uses the same physical factorization so that sensed and commanded interaction are represented consistently.
Using the online friction estimate \(\hat{\mu}_t\) and contact activation \(\alpha_t\in[0,1]\), we define
\begin{equation}
\tilde{\bm{w}}_t
=
\left(
\alpha_t\tilde f_{\mathrm{nor},t},
\alpha_t\rho_t(\hat{\mu}_t),
\alpha_t
\right).
\label{eq:interaction_observation}
\end{equation}
When \(f_{\mathrm{nor},t}\leq f_{\epsilon}\), the tangential component is set to zero.
The activation suppresses interaction channels outside contact while explicitly indicating the current contact state.
Unlike the action coordinate, the observed tangential residual remains linear and unclipped so that measured variations in the interaction are retained.
The contact activation \(\alpha_t\) and friction estimate \(\hat{\mu}_t\) are obtained online as described in Section~\ref{sec:method:estimation}.

%Together, these choices factor out three sources of variation that are incidental to the demonstrated behavior. Relative poses remove dependence on global configuration, the contact normal removes dependence on local surface orientation, and the friction scale removes the dominant contact-dependent scaling of tangential force. The policy can therefore learn motion and interaction in coordinates that remain consistent across changing contact conditions, while the contact factor can be estimated independently and reused to realize the same behavior on previously unseen surfaces and objects.

Relative poses remove global-frame dependence, while the contact normal and friction scale account for local surface orientation and the dominant scaling of tangential force.

\subsection{Policy Learning}
\label{sec:method:policy}

The policy learns to reproduce the demonstrated motion and interaction in contact-factorized coordinates, decoupling the learned behavior from the contact factors governing its physical realization, as illustrated in Fig.~\ref{fig:system}.
To construct consistent interaction targets, we associate each training contact condition \(\kappa\), defined by its material and contact configuration, with a nominal friction reference \(\bar{\mu}_{\kappa}\).
We compute \(\bar{\mu}_{\kappa}\) as the median tangential-to-normal force ratio over valid sliding samples and keep it fixed throughout each demonstration.
This removes the nominal friction scale while preserving temporal variation in the demonstrated interaction.
Accordingly, the interaction targets are
\(c_t^{\mathrm{GT}}=c_t(\bar{\mu}_{\kappa})\) and
\(\tilde f_{\mathrm{nor},t}^{\mathrm{GT}}\) from (\ref{eq:normal_force_normalization}).
Normal loading is supervised over the full trajectory, including zero loading in free space.

Interaction observations instead use the causal friction estimate \(\hat{\mu}_t\) available at deployment through (\ref{eq:interaction_observation}).
Using this time-varying estimate to construct action targets could absorb part of the demonstrated interaction variation into the estimated contact factor.
We therefore use the fixed reference \(\bar{\mu}_{\kappa}\) to define the interaction behavior to be learned, while \(\hat{\mu}_t\) is reserved for causal observation and execution adaptation.

For a prediction horizon \(H\), the demonstrated action sequence is
\(
\bm{y}^{1}
=
\{\bm{a}^{\mathrm{GT}}_{t+h}\}_{h=0}^{H-1},
\)
where each action combines the demonstrated relative motion with the corresponding factorized interaction target.
We train a conditional flow-matching policy \(v_{\theta}\) to generate this sequence from the current policy observation.
Following standard flow matching~\cite{lipman2022flow}, we sample
\(\bm{y}^{0}\sim\mathcal{N}(\bm{0},\bm{I})\) and interpolate between \(\bm{y}^{0}\) and \(\bm{y}^{1}\), with target velocity \(\bm{y}^{1}-\bm{y}^{0}\).
Let \(\bm{e}_{t,h}^{\mathrm{pose}}\), \(e_{t,h}^{\tilde f}\), and \(e_{t,h}^{c}\) denote the pose, normal-loading, and tangential-interaction components of the flow prediction error.
We optimize
\begin{equation}
\mathcal{L}(\theta)=\mathbb{E}\!\Bigl[
\tfrac{1}{H}\sum_{h=0}^{H-1}\!\bigl(
\|\bm{e}_{t,h}^{\mathrm{pose}}\|_2^2
+\lambda_{\mathrm{int}}\bigl(
|e_{t,h}^{\tilde f}|^2+\alpha_{t+h}|e_{t,h}^{c}|^2
\bigr)\bigr)\Bigr].
\end{equation}
Pose and normal loading are supervised throughout the trajectory, while tangential interaction is supervised only during valid contact.
The tangential term is evaluated when \(f_{\mathrm{nor},t+h}>f_{\epsilon}\), and padded steps at episode boundaries are excluded.

\subsection{Online Contact Estimation \& Execution Adaptation}
\label{sec:method:estimation}
At deployment, we estimate \(z_t=(\bm n_t,\mu_t)\) online to convert outputs of a fixed policy into motion and force references for the hybrid controller under unseen contact conditions. We first estimate the contact state and local geometry.
The contact activation \(\alpha_t \in [0,1]\) is computed by applying a smoothstep function to the measured force magnitude between prescribed lower and upper thresholds.
Because force magnitude is independent of the contact frame, \(\alpha_t\) can be obtained before estimating the local contact normal.
The local normal \(\hat{\bm n}_t\) is estimated from recent pose, rotation relative to the initial state, and wrench history using a separately trained estimator.
The estimator is trained using the same demonstrations as the task policy, with an activation-weighted cosine loss against normals obtained from registered training geometry.
This unit normal then defines the normal direction and tangent plane used for friction estimation.

During active contact with \(f_{\mathrm{nor},t}>f_{\epsilon}\) and sufficient tangential motion,
\(
\mu_t^{\mathrm{meas}}
=
\frac{\|\bm f_{\mathrm{tan},t}\|}{f_{\mathrm{nor},t}}
\approx \mu_t.
\)
A scalar Kalman filter tracks \(\mu_t\) and provides the online estimate \(\hat{\mu}_t\), initialized from a condition-independent prior \(\mu_0\).

Given the estimated contact factor, the predicted interaction is converted into physical force references,\\
\begin{equation}
\begin{aligned}
f_{\mathrm{nor},t}^{\mathrm{des}} &= f_{\mathrm{nor},\max} \Pi_{[0,1]} \left( \tilde f_{\mathrm{nor},t}^{\mathrm{des}} \right), \\
f_{\mathrm{tan},t}^{\mathrm{des}} &= \hat{\mu}_t f_{\mathrm{nor},t}^{\mathrm{des}} 
\Pi_{[\rho_{\min},\rho_{\max}]} \left( e^{c_t^{\mathrm{des}}} \right).
\end{aligned}
\label{eq:force_reparameterization}
\end{equation}
Here, \(\hat{\mu}_t\) restores the friction-dependent scale of the tangential interaction.

The estimated normal similarly adapts the predicted motion to the local contact geometry,
\begin{equation}
\begin{aligned}
\Delta\bm x_{\mathrm{tan},t} &= \left( \bm I-\hat{\bm n}_t\hat{\bm n}_t^\top \right) \Delta\bm x_t, \\
g_t &= \operatorname{smoothstep} \left( \frac{ f_{\mathrm{nor},t}^{\mathrm{des}}-f_{g,0} }{ f_{g,1}-f_{g,0} } \right),\\
\Delta\bm x_t^{\mathrm{des}} &= (1-g_t)\Delta\bm x_t + g_t\Delta\bm x_{\mathrm{tan},t}.
\end{aligned}
\label{eq:contact_pose}
\end{equation}
Here, \(f_{g,0}\) and \(f_{g,1}\) are the lower and upper normal-loading thresholds for blending the original and tangential motions.
As the desired normal loading approaches zero, the original translational motion is recovered.
The relative orientation is preserved as \(\Delta\bm R_t^{\mathrm{des}}=\Delta\bm R_t\). The desired tangential force is applied as a feedforward term, while normal force is regulated through feedback along the estimated contact normal. The resulting motion and force references are executed by a velocity-based hybrid position-force controller~\cite{raibert1981hybrid}.
Each committed action is reparameterized using the latest \(\hat{\bm n}_t\) and \(\hat{\mu}_t\), enabling continuous adaptation without additional policy inference within the committed action chunk.

\begin{figure*}[t]
    \centering
    \vspace{4pt}
    \includegraphics[width=2.0\columnwidth]{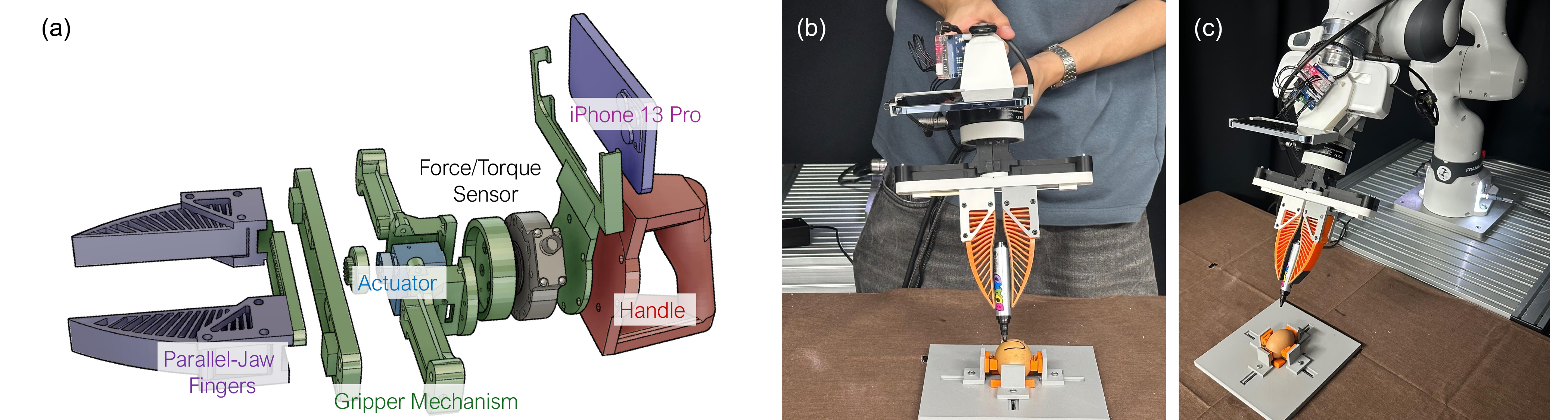}
    \caption{
    \textbf{Hardware setup.}
    (\textbf{a}) The handheld demonstration interface integrates an actuated parallel-jaw gripper, a wrist-mounted force/torque sensor, and an iPhone for visual-inertial tracking.
    (\textbf{b}) During data collection, the human demonstrator directly operates the interface, enabling intuitive demonstration of contact-rich interactions while synchronously recording motion, visual observations, and contact forces.
    (\textbf{c}) During robot deployment, the handle is removed and the same gripper and sensing assembly is mounted on the robot, reducing end-effector mismatch between human demonstration and execution.}
    \label{fig:hardware}
\end{figure*}

\begin{figure}[t]
    \centering
    \vspace{4pt}
    \includegraphics[width=1.0\columnwidth]{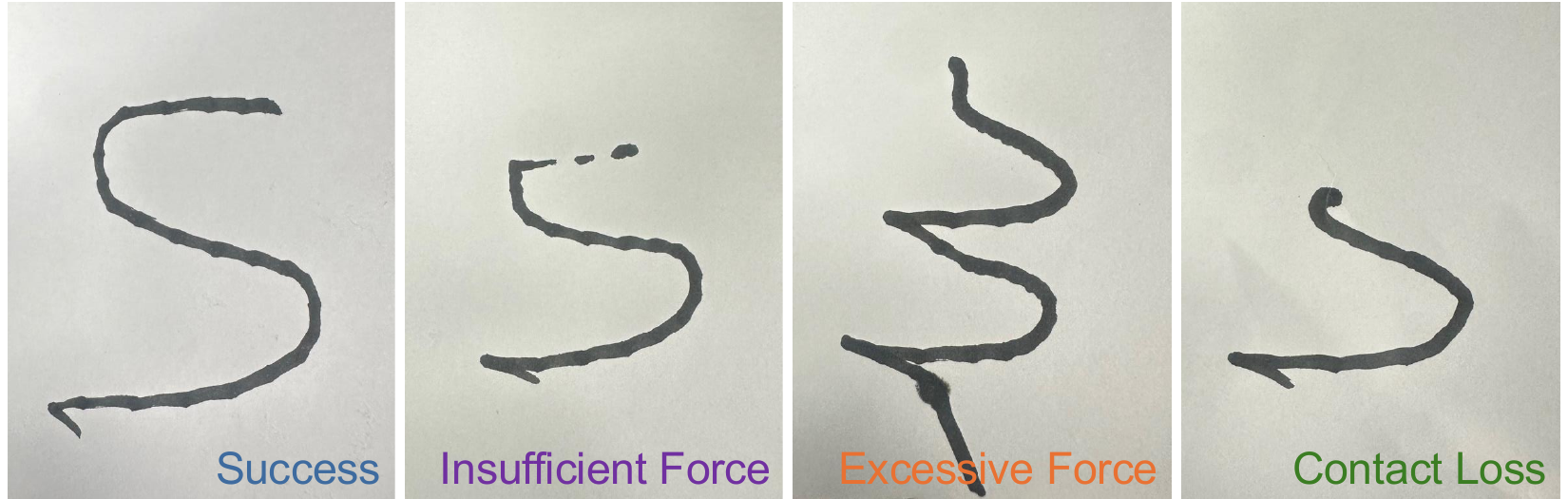}
    \caption{
    \textbf{Success and failure modes in the marking task.}
    Representative outcomes of the marking task. A successful trial produces a continuous target trajectory, while failures arise from insufficient contact force, excessive contact force, or loss of contact during execution.
    }

    \label{fig:failure}
\end{figure}

\begin{figure*}[t]
    \centering
    \vspace{4pt}
    \includegraphics[width=2.0\columnwidth]{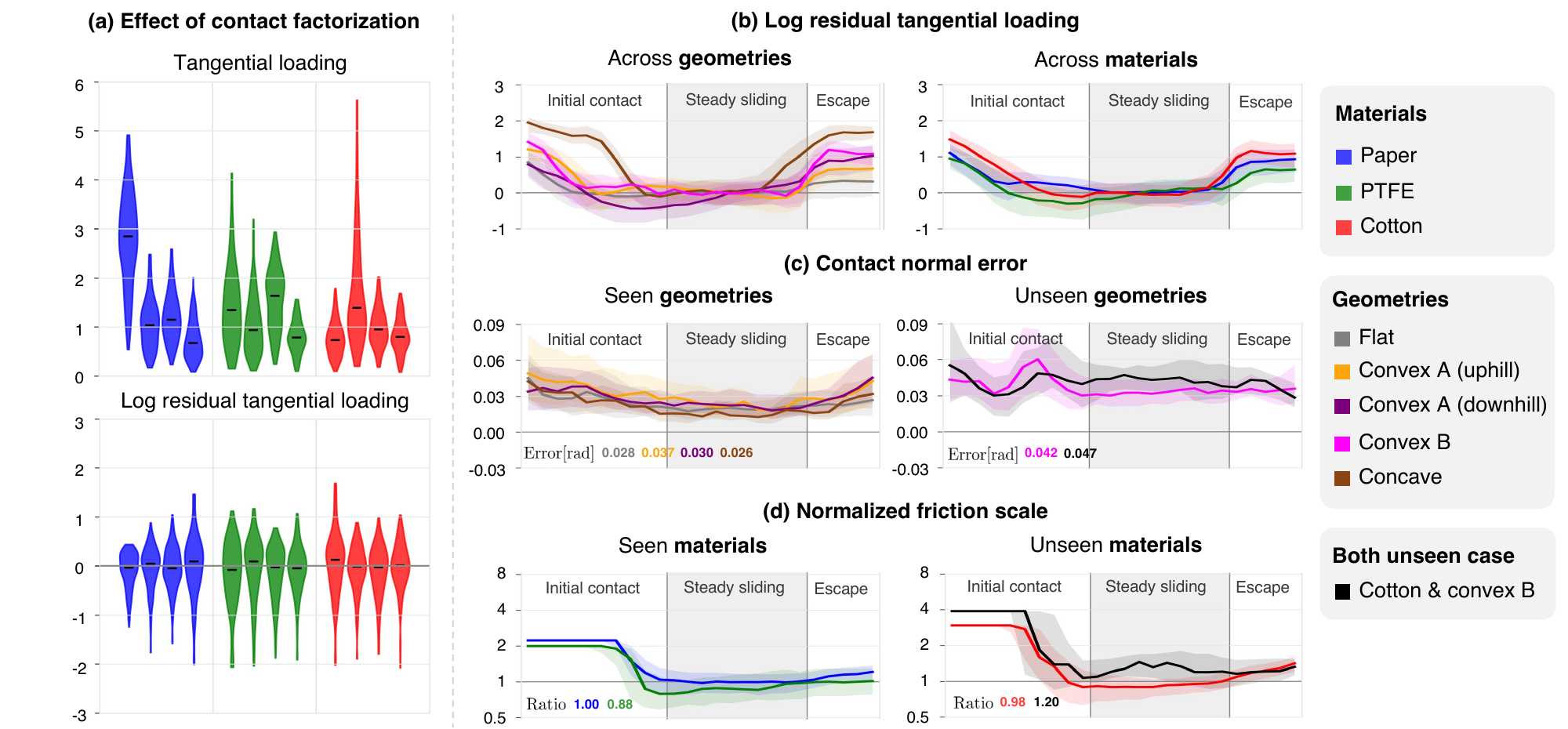}
    \caption{
\textbf{Contact factorization and online contact factor estimation.}
(\textbf{a}) Tangential loading $\lvert f_{\mathrm{tan}}\rvert$ varies across contact conditions, while the factorized coordinate $c=\log\rho$, with $\rho=f_{\mathrm{tan}}/(\bar{\mu}f_{\mathrm{nor}})$, remains centered near zero during steady sliding. For each material, the four violins correspond, from left to right, to flat, convex A (uphill), convex A (downhill), and convex B.
(\textbf{b}) The residual $c$ exhibits a consistent temporal structure across geometries and materials, with larger deviations during initial contact and escape.
(\textbf{c}) Online contact-normal estimation remains accurate on both seen and unseen geometries.
(\textbf{d}) The normalized friction estimate $\hat{\mu}/\bar{\mu}$ converges toward unity during steady sliding for seen and unseen materials.
Cotton and convex B are excluded from estimator training; lines show per-condition medians and shaded regions indicate interquartile ranges.
}
    \label{fig:contact}
\end{figure*}

\section{Experiments}
\label{sec:experiments}

The proposed method is then evaluated on a real-robot drawing task using an FR3 manipulator (Franka Robotics GmbH)~\cite{haddadin2022franka}.
The evaluation considers unseen objects with different contact geometries and frictional properties to test generalization under changing physical contact conditions.
Demonstrations are collected with the handheld tool, providing an intuitive interface for acquiring contact-rich behavior while allowing force-domain interaction to be learned despite differences in physical realization between demonstration and robot execution.
Our experiments address three research questions:
\begin{enumerate}
    \item Can the contact factor be estimated reliably on unseen objects and surfaces?
    \item Can contact factorization effectively decouple learned behavior from contact-dependent physical realization?
    \item Does this factorization improve object generalization across changes in contact geometry and frictional properties, and how does each component contribute?
\end{enumerate}

\begin{figure}[t]
    \centering
    \vspace{4pt}
    \includegraphics[width=1.0\columnwidth]{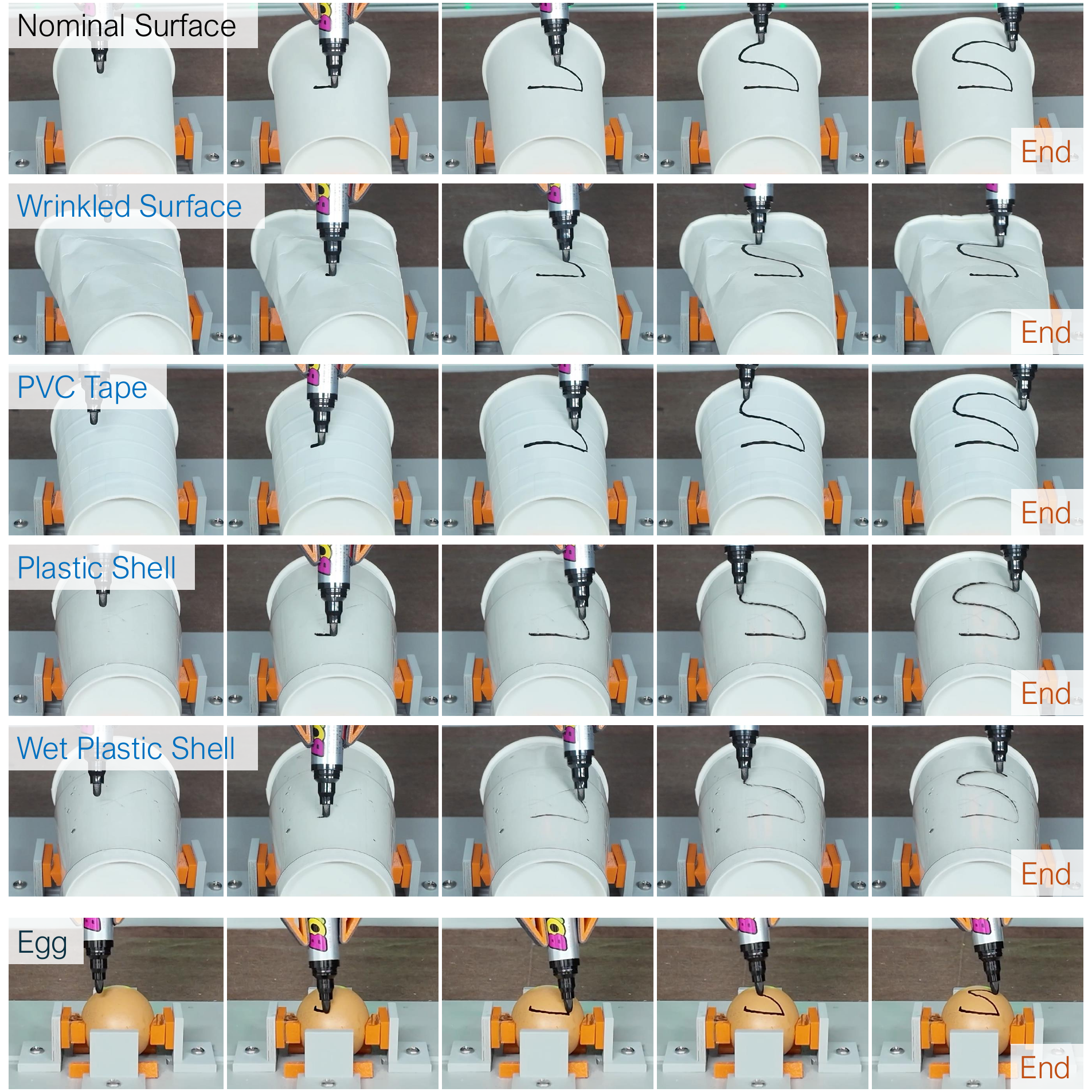}
    \caption{
    \textbf{Timelapse of deploying \ourmethod{} across unseen objects.}
    Despite substantial variations in surface geometry, material properties, and contact conditions, \ourmethod{} adapts the interaction online and consistently executes the intended marking trajectory, demonstrating robust generalization beyond the conditions observed during training.
    }
    \label{fig:timelapse}
\end{figure}

\subsection{Experimental Setup}
\label{sec:exp}

We design the following task to enable systematic comparison with the baselines and isolate the contributions of individual components under controlled contact variations. Additional contact-rich manipulation tasks demonstrating broader applicability are provided on the project page.

\paragraph{Task Design}
We consider controlled contact interactions and a real-robot target task of marking an ``S'' on fragile objects.
We use two objects with complementary physical properties:
\begin{enumerate}
\item \textbf{Paper Cup:} a lightweight and compliant object that can deform under small contact forces. We vary its geometry using pre-deformed, folded, or wrinkled cups.
\item \textbf{Egg:} a rigid and brittle object with a continuously curved surface. We vary its geometry using different-sized eggs.
\end{enumerate}
For the paper cup, we additionally vary the contact surface to introduce different frictional conditions:
\begin{enumerate}
\item \textbf{PVC Tape:} the contact region is covered with PVC tape to create a relatively higher-friction surface.
\item \textbf{Plastic Shell:} a smooth plastic layer is placed over the cup to create a lower-friction contact surface.
\item \textbf{Wet Plastic Shell:} the plastic surface is wetted to further reduce friction and create a more slippery contact condition.
\end{enumerate}
Demonstrations are collected on a nominal instance of each object, while these geometric and surface variations are introduced only at deployment.
The marking task requires sustained sliding contact under limited force while continuously adapting to changes in local surface orientation and contact conditions.

\paragraph{Data Collection}

Demonstrations are collected using the handheld sensing interface in Fig.~\ref{fig:hardware}.
The device integrates an actuated parallel-jaw gripper, a wrist-mounted six-axis force/torque sensor (AFT200, Aidin Robotics), and an iPhone 13 Pro providing RGB images and six-degree-of-freedom (DoF) pose estimation through ARKit (Apple Inc.).
Electronic gripper actuation prevents the operator's grasping force from contaminating the wrist wrench.
RGB, pose, and gripper state are recorded at 30\,Hz, while the wrist wrench is sampled at 500\,Hz.
All measurements are transformed into a calibrated tool frame using fixed extrinsics, and the wrench is bias-corrected and gravity-compensated using the onboard IMU.

\paragraph{Robot Deployment}

The handheld tool is mounted to the robot through a dedicated adapter, preserving the sensing configuration and tool-frame convention used during demonstration.
For policy inference, wrench measurements are aligned with the 30\,Hz RGB and pose streams, and pose trajectories are converted to the relative representation in Section~\ref{sec:method:representation}.
The 500\,Hz wrench stream is retained for online contact processing and contact-factor estimation as described in Section~\ref{sec:method:estimation}.
The same calibration and wrench preprocessing are used during demonstration and deployment.

\paragraph{Evaluation Protocol and Metrics}
The contact-normal estimator is trained using the same demonstrations as the corresponding policy.
For each object category, all methods are trained on the same set of 200 human demonstrations collected using a nominal instance and evaluated under unseen deployment conditions. 
For surface variations, we evaluate three unseen conditions on paper cups, with 10 trials per method per condition.
For geometric variations, we evaluate four unseen deformation conditions on paper cups, with 5 trials per method per condition, and five unseen egg instances, with 5 trials per method per instance. 
We report the task success rate as the primary metric. 
To further characterize unsuccessful trials, we categorize failures into four modes.
The three contact-related failure modes are illustrated in Fig.~\ref{fig:failure}.
We additionally consider failures caused by incorrect motions generated by the visuomotor policy:
\begin{itemize}
\item \textbf{Insufficient Force.} The marker remains in contact with the object, but the applied force is too small to produce a consistent visible mark, resulting in faint or incomplete trajectories.
\item \textbf{Excessive Force.} The applied force is too large, causing substantial deformation or damage to the fragile object during marking.
\item \textbf{Contact Loss.} The marker loses contact with the object surface during execution, producing gaps or discontinuities in the marked trajectory.
\item \textbf{Incorrect Motion.} The visuomotor policy generates motion that substantially deviates from the intended marking trajectory, preventing successful task completion.
\end{itemize}

\subsection{Surface Estimation and Interaction Representation}
\label{ref:adaptation}

We first evaluate the adaptation module in a separate controlled-sliding experiment, distinct from the task-level evaluation in Sec~\ref{sec:quant}.
These experiments assess whether contact-dependent factors can be accurately estimated under seen and unseen surface conditions and whether the proposed interaction representation separates contact-dependent force scaling while retaining the temporal structure of measured interactions.

We collect controlled sliding interactions with line contact on a small set of surface geometries and materials.
The contact normal estimator is trained using paper and PTFE over flat, convex uphill, convex downhill, and concave uphill configurations, with 100 training episodes per material and configuration combination.
The contact-factor estimator first computes force-based contact activation and estimates the contact normal from motion and wrench histories using a GRU.
The estimated normal is then used to decompose the measured force and update the friction estimate with a scalar Kalman filter during steady sliding.
Additional implementation and data-collection details are provided on the project page.
For evaluation, we consider both seen conditions and held-out conditions containing unseen geometries, unseen materials, or both.
Cotton serves as the unseen material, and an unseen convex geometry is evaluated with paper, PTFE, and cotton. More details are available on the project page.

Fig.~\ref{fig:contact} summarizes the results.
Raw tangential loading varies across materials and geometries, whereas factorized residuals are centered near zero during steady sliding (a).
The residuals retain phase-dependent variation, and departures from $\rho=1$ outside steady sliding support learning $\rho$ rather than fixing it to unity (b).
Contact-normal errors average $0.030$ rad on seen geometries and 0.045 rad on unseen geometries (c).
Initialized from a condition-independent prior, friction estimates approach their reference scales during steady sliding (d).
With cotton and convex B both unseen, the normal error is $0.047$ rad and the steady-sliding median friction ratio is $1.20$.
These results validate the contact factorization and demonstrate that the adaptation module can estimate contact factors online under unseen contact conditions (Research Question 1).

\begin{table*}[t]
\centering
\vspace{10pt}
\scriptsize
\caption{
\textbf{Quantitative results across contact geometry variations.}
We vary the local surface geometry using folds and wrinkles with different orientations (horizontal or vertical) and deformation severities (mild or severe) across paper cups and eggs.
Success rates and failure-mode breakdowns are reported as percentages over 5 trials for each method and condition.
Higher success rates are better.
Failure rates are shown in parentheses in the following order: Insufficient Force (purple), Excessive Force (orange), Contact Loss (green), and Incorrect Motion (red); lower values are better.
}
\label{tab:geometry}

\setlength{\tabcolsep}{3.0pt}
\renewcommand{\arraystretch}{1.2}

\resizebox{2.05\columnwidth}{!}{%
\begin{tabular}{l ccccc|c}
\toprule

\multirow{2}{*}{Method}
& \multicolumn{5}{c|}{Paper Cup}
& \multirow{2}{*}{Egg} \\

\cmidrule(lr){2-6}

& Normal
& Horizontal (mild)
& Horizontal (severe)
& Vertical (mild)
& Vertical (severe)
&
\\

\midrule

Visual Observation Only
& 0\% \failrates{0}{80}{20}{0}
& 0\% \failrates{0}{0}{100}{0}
& 0\% \failrates{40}{0}{60}{0}
& 0\% \failrates{40}{0}{60}{0}
& 0\% \failrates{20}{20}{60}{0}
& 0\% \failrates{0}{0}{96}{4}
\\

Force as Observation
& 0\% \failrates{80}{0}{20}{0}
& 20\% \failrates{40}{0}{20}{20}
& 0\% \failrates{0}{0}{100}{0}
& 0\% \failrates{100}{0}{0}{0}
& 0\% \failrates{100}{0}{0}{0}
& 0\% \failrates{0}{0}{0}{100}
\\

No Force-Factorization
& 80\% \failrates{0}{0}{0}{20}
& 60\% \failrates{40}{0}{0}{0}
& 0\% \failrates{0}{0}{100}{0}
& \textbf{80\%} \failrates{20}{0}{0}{0}
& 40\% \failrates{40}{0}{20}{0}
& 0\% \failrates{0}{0}{0}{100}
\\

No Learned Contact-Normal Estimation
& 40\% \failrates{40}{0}{20}{0}
& 0\% \failrates{100}{0}{0}{0}
& 0\% \failrates{100}{0}{0}{0}
& 0\% \failrates{0}{0}{100}{0}
& 0\% \failrates{0}{0}{100}{0}
& 4\% \failrates{0}{56}{0}{40}
\\

\rowcolor{gray!15}
\textbf{Full Method}
& \textbf{100\%} \failrates{0}{0}{0}{0}
& \textbf{80\%} \failrates{0}{0}{20}{0}
& \textbf{60\%} \failrates{0}{0}{40}{0}
& 60\% \failrates{40}{0}{0}{0}
& \textbf{60\%} \failrates{0}{0}{40}{0}
& \textbf{68\%} \failrates{24}{8}{0}{0}
\\

\bottomrule
\end{tabular}%
}
\vspace{6pt}
\end{table*}

\begin{table*}[t]
\centering
\small
\caption{
\textbf{Quantitative results across contact surface variations.}
We vary the contact material from the nominal surface (Normal) to PVC tape and a smooth plastic shell, with the wet plastic shell introducing an additional low-friction surface condition.
Success rates and failure-mode breakdowns are reported as percentages over 10 trials for each method and condition.
Higher success rates are better.
Failure rates are shown in parentheses in the following order: Insufficient Force (purple), Excessive Force (orange), Contact Loss (green), and Incorrect Motion (red); lower values are better.}
\label{tab:material}

\setlength{\tabcolsep}{3.0pt}
\renewcommand{\arraystretch}{1.2}

\makeatletter\def\@captype{table*}
\resizebox{1.6\columnwidth}{!}
{
\begin{tabular}{l cccc}
\toprule

\textbf{Method}
& \textbf{Normal}
& \textbf{PVC Tape}
& \textbf{Plastic Shell}
& \textbf{Wet Plastic Shell}
\\

\midrule

Visual Observation Only
& 0\% \failrates{0}{80}{20}{0}
& 0\% \failrates{50}{10}{40}{0}
& 10\% \failrates{30}{0}{60}{0}
& 20\% \failrates{10}{0}{70}{0}
\\

Force as Observation
& 0\% \failrates{80}{0}{20}{0}
& 0\% \failrates{30}{0}{0}{70}
& 10\% \failrates{90}{0}{0}{0}
& 0\% \failrates{100}{0}{0}{0}
\\

No Force-Factorization
& 80\% \failrates{0}{0}{0}{20}
& 40\% \failrates{0}{60}{0}{0}
& 50\% \failrates{0}{50}{0}{0}
& 60\% \failrates{0}{0}{0}{40}
\\

No Learned Contact-Normal Estimation
& 30\% \failrates{50}{0}{20}{0}
& 30\% \failrates{30}{10}{10}{20}
& 10\% \failrates{20}{0}{70}{0}
& 10\% \failrates{20}{0}{70}{0}
\\

\rowcolor{gray!15}
\textbf{Full Method}
& \textbf{100\%} \failrates{0}{0}{0}{0}
& \textbf{100\%} \failrates{0}{0}{0}{0}
& \textbf{100\%} \failrates{0}{0}{0}{0}
& \textbf{90\%} \failrates{10}{0}{0}{0}
\\

\bottomrule
\end{tabular}%
}
\end{table*}

% \begin{table*}[t]
% \centering
% \begin{tabular*}{\textwidth}{
% @{\extracolsep{\fill}}
% l
% c
% ccccc
% @{}
% }
% \hline
% \multirow{2}{*}{Method}
% & \multirow{2}{*}{\shortstack{Success [\%]}}
% & \multicolumn{5}{c}{Failure Modes} \\
% \cline{3-7}
% & & A & B & C & D & E \\
% \hline

% UMI
% & 24/30 (80.0\%)
% & 1 & 0 & 0 & 0 & 5 \\

% ForceMimic
% & 23/30 (76.7\%)
% & 2 & 0 & 5 & 0 & 0 \\

% \hline

% Force as obs.
% & 23/30 (76.7\%)
% & 6 & 0 & 0 & 0 & 1 \\

% Ours (w/o $\hat{\bm n}$)
% & 11/30 (36.7\%)
% & 6 & 2 & 6 & 2 & 3 \\

% Ours
% & 27/30 (90.0\%)
% & 3 & 0 & 0 & 0 & 0 \\

% \hline
% \end{tabular*}
% \vspace{-4mm}
% \end{table*}

\subsection{Quantitative Evaluation}
\label{sec:quant}

We evaluate whether our method enables robust transfer from demonstrations collected under nominal conditions to robot deployment under varying physical conditions, as presented in Fig.~\ref{fig:timelapse}. 
To characterize the contribution of its individual design choices, we jointly compare against prior approaches adapted to our setting and controlled variants of our method, where individual components are modified or removed:

\begin{itemize}
\item \textbf{Visual Observation Only:}
We design a vision-based imitation learning baseline that predicts end-effector motion without explicitly representing physical interaction. Inspired by UMI~\cite{chi2024umi}, a representative approach in this category, the baseline retains our flow-matching architecture but removes the contact-adaptation encoder and decoder. This controlled comparison evaluates the benefit of explicitly modeling physical interaction under varying contact conditions.

\item \textbf{Force as Observation:}
This variant provides the raw measured interaction force directly to the visuomotor policy as an additional observation, without contact factorization.
Comparison with the full method evaluates whether factorizing interaction from contact-dependent effects provides benefits beyond simply conditioning the policy on force feedback.

\item \textbf{No Force-Factorization:}
Inspired by the direct force prediction in ForceMimic~\cite{liu2025forcemimic}, this variant represents force observations and actions in physical units rather than factorized coordinates.
For a controlled comparison, we retain our policy backbone and hybrid execution, constructing force directions from the estimated contact normal and predicted motion rather than directly predicting force vectors.
Comparison with the full method evaluates the benefit of contact factorization in separating learned interaction from its contact-dependent physical execution.

\item \textbf{No Learned Contact-Normal Estimation:}
This variant replaces the learned contact-normal estimator with an analytic estimator based on measured motion and force, while retaining the same policy and remaining components.
Comparison with the full method evaluates the role of learned contact-normal estimation in transferring across varying surface geometries.
\end{itemize}

Table~\ref{tab:geometry} shows that FACE achieves 65\% success (13/20) across unseen paper cup deformations, compared with 0\% for the analytic contact-normal variant.
On unseen eggs, the corresponding rates are 68\% and 4\%.
Excessive-force failures on eggs decrease from 56\% with the analytic estimator to 8\% with FACE, although insufficient-force failures remain at 24\%.
Visual Observation Only exhibits contact loss in 96\% of egg trials, whereas no such failures are observed with FACE.
No Force-Factorization achieves 45\% across cup deformations and exceeds FACE in Vertical (mild) by one trial.
These results highlight the contribution of learned contact-normal estimation, although gains vary across conditions.

Table~\ref{tab:material} shows that FACE succeeds in 29/30 trials (96.7\%) across the three unseen surface conditions, compared with 15/30 (50\%) for No Force-Factorization.
Of its 15 failures, 11 are excessive-force failures on PVC tape and the plastic shell, while 4 are incorrect-motion failures on the wet plastic shell.
The excessive-force failures are consistent with the sensitivity of direct physical force targets to changes in contact-dependent force scale, whereas FACE's only failure is insufficient force on the wet plastic shell.
The analytic contact-normal variant succeeds in 5/30 trials, further supporting the role of learned contact-normal estimation in translating factorized interaction into physical commands.
Visual Observation Only and Force as Observation succeed in 3/30 and 1/30 trials, respectively, indicating that neither omitting physical interaction nor simply conditioning on raw force is sufficient under surface changes.

Together, these results support separating learned interaction from its contact-dependent physical execution to improve task generalization under unseen geometry and surface conditions (Research Question 2).
The ablations highlight contributions from both force factorization and learned contact-normal estimation, while remaining insufficient-force and contact-loss failures indicate limitations in contact regulation (Research Question 3).
\section{Conclusion}
We present \ourmethod{}, a contact-factorized imitation learning framework that separates intended task behavior from environment-dependent contact factors.
Force observations and actions are represented in normalized coordinates defined by the local contact frame and friction scale.
A learned contact-normal estimator and an online friction filter enable execution to adapt to current contact conditions without updating the task policy.
Using demonstrations collected with a force-sensing handheld interface, our real-robot experiments show improved transfer to unseen surface properties and geometries.
Controlled comparisons highlight the complementary roles of force factorization and contact-normal adaptation in maintaining the intended interaction.
These results support separating task behavior from contact-dependent force scale and direction as a means of improving generalization in contact-rich imitation learning.
As a potential limitation, friction estimation remains challenging when sliding is limited. Integrating visual cues provides a concrete path toward resolving this partial observability and extending the framework to deformable and multi-contact interactions.

% {
% \footnotesize
% \noindent
% {\bf Acknowledgements}
% }

{
{
\vspace{12pt}
\small
\noindent
{\bf Acknowledgements}
This work was supported by the Ministry of Trade, Industry and Energy (MOTIE) (RS-2024-00419641, RS-2025-25449039) and the Ministry of Science and ICT through the Korea Basic Science Institute (RS-2025-00564593), Republic of Korea.
}

% only do this for icra
% \renewcommand{\baselinestretch}{0.985} 
% \renewcommand*{\bibfont}{\small}
% \printbibliography
% }

% % only do this for arxiv
\renewcommand{\baselinestretch}{1.05} 
\renewcommand*{\bibfont}{\scriptsize}
\printbibliography
}

\end{document}